\documentclass[a4paper,11pt]{article}
\usepackage{gensymb}
\usepackage{fourier}
\usepackage{color}
 \usepackage{graphicx}
\usepackage{url}
\usepackage[affil-it]{authblk}
\usepackage{amsmath}
\usepackage{wrapfig}
\usepackage{lipsum,booktabs}
\usepackage{enumitem}
\usepackage[T1]{fontenc}
\usepackage{times}
\usepackage{multirow}
\usepackage[justification=centering]{caption}
\usepackage{cite}

\begin{document}

\title{Texture segmentation with Fully Convolutional Networks}

\author{Vincent Andrearczyk \& Paul F. Whelan}
\affil{Vision Systems Group, School of Electronic Engineering, Dublin City University, Glasnevin, Dublin 9, Ireland}
\date{}
\maketitle
\thispagestyle{empty}

\begin{abstract}
In the last decade, deep learning has contributed to advances in a wide range computer vision tasks including texture analysis.
This paper explores a new approach for texture segmentation using deep convolutional neural networks, sharing important ideas with classic filter bank based
texture segmentation methods.
Several methods are developed to train Fully Convolutional Networks to segment textures in various applications.
We show in particular that these networks can learn to recognize and segment a type of texture, e.g. wood and grass from texture recognition datasets (no training segmentation).
We demonstrate that Fully Convolutional Networks can learn from repetitive patterns to segment a particular texture from a single image or even a part of an image.
We take advantage of these findings to develop a method that is evaluated on a series of supervised and unsupervised experiments and improve the state of the
art on the Prague texture segmentation datasets.

\end{abstract}

\let\thefootnote\relax\footnote{\em Article under consideration at Pattern Recognition Letters.}
\section{Introduction}
The analysis of texture is, together with color, a key step in many computer vision tasks.
Texture regions are generally defined as the statistical spatial distribution of their pixel intensities and can be described,
among others, as fine, coarse, rippled or irregular \cite{haralick1973textural}.
This paper focuses on the segmentation of images into multiple texture regions.
Real world images exhibit various textures at different scales and frequencies which makes its analysis challenging.
Many classic texture segmentation methods involve extracting the response to hand-crafted filter banks followed by a clustering or
classification method \cite{liu2006image,mevenkamp2016variational,yuan2015factorization,shi2000normalized,paragios2002geodesic}.
Convolutional Neural Networks (CNNs) are well suited for texture analysis \cite{andrearczyk2016using,andrearczyk2016deep,cimpoi2016deep} with a design similar
to filter banks trained with powerful learning algorithms.
They obtain excellent results on texture recognition datasets by discarding the overall shape analysis of classic network architectures.
CNNs share several ideas with classic texture segmentation methods including filter banks and
multi-scale analysis (e.g. skip layers \cite{long2015fully}, hypercolumns \cite{hariharan2015hypercolumns}) and classification in the feature space.
A Fully Convolutional Network (FCN) \cite{long2015fully} is a type of CNN which can efficiently perform semantic segmentation of
objects (e.g. a car, an animal or a person).
This paper thus investigates the power of FCNs in the segmentation of texture images.
We develop a fully convolutional architecture optimized for the segmentation of textures which discards very deep shape information and
combines the response to filter banks at various depths to make use of local (shallow features) and more global and
abstract information (deeper features).

This paper differentiates between two major texture segmentation problems with different application focus.
The first task is the semantic segmentation of types of textures with a potentially large intra-class variation (e.g. wood, grass and textile).
This first problem is relatively similar to a classic image classification or segmentation task in which an algorithm is trained to detect an
object with many possible variations (shape, scale, lighting etc.) in a supervised manner with many samples of each class.
The second problem is to learn to segment textures with a single small training (supervised) sample or with no training data at all (unsupervised).
Note that the approach is unsupervised but the training of FCN itself is supervised as explained in Section \ref{sec:methodDescription}.
For this task in particular, we make use of the repetitivity of the textural patterns which can be efficiently learned by finetuning a FCN with little training data.
Three sets of experiments are therefore developed to evaluate our approaches on these tasks including
A) a supervised semantic segmentation with multiple training images, B) a supervised segmentation with a single training image per class and C) an unsupervised segmentation.
We evaluate our approach on images that we have generated (experiment A) as well as on the Prague texture segmentation benchmark \cite{haindl2008texture} (experiments B and C),
on which we considerably improve the state of the art.

The contributions of this paper include
i) a FCN architecture for texture segmentation combining the analysis of simple local texture patterns
with more global and complex patterns while discarding the overall shape analysis;
ii) we demonstrate the potential of such network to learn to segment textures with non-segmented texture images;
iii) we show that this network can be trained with very little training data
(single image per class and even parts of the test image itself) which enables
learning texture segmentation in an unsupervised framework.

The rest of this paper is organized as follows. The related work is reviewed in Section \ref{sec:relatedWork}.
Our approach including network architecture and training methods is described in Section \ref{sec:methodDescription}.
Three sets of experiments are presented in Sections \ref{subsec:experimentA}, \ref{subsec:experimentB} and \ref{subsec:experimentC} including supervised and unsupervised
texture segmentation.

\section{Related work}\label{sec:relatedWork}
Texture analysis often involves the extraction of handcrafted local features combined with a learning algorithm to perform
the desired task in the feature domain such as classification \cite{randen1999filtering} or clustering \cite{jain1991unsupervised,liu2006image,mevenkamp2016variational,yuan2015factorization}.
In particular, a wide range of texture segmentation methods involve the extraction of texture descriptors
based on local responses to hand-designed filter banks typically with a set of scales and orientations.
Gabor filters are the most commonly used filters in texture segmentation \cite{jain1991unsupervised,paragios2002geodesic,liu2006image,mevenkamp2016variational,yuan2015factorization}.
The response to such filter banks is often used to obtain powerful texture descriptors,
namely local spectral histograms \cite{liu2006image,mevenkamp2016variational,yuan2015factorization}.
Other filters include intensity, gradients and Laplacian filters \cite{liu2006image}, difference of oriented Gaussian filters \cite{shi2000normalized},
as well as isotropic and anisotropic filters \cite{paragios2002geodesic}.
Other types of popular texture features used in texture segmentation include statistical descriptors
such as Local Binary Patterns (LBPs) \cite{ojala1999unsupervised} and co-occurrence matrices \cite{chen1979segmentation}
as well as wavelet transforms \cite{arivazhagan2003texture}.
Segmentation of texture descriptors can be performed, among others by graph cut \cite{shi2000normalized},
curve evolution with level-set optimization \cite{paragios2002geodesic}, region growing and merging \cite{chen1979segmentation} and
functional minimization (Mumford-Shah functional) \cite{liu2006image,mevenkamp2016variational}.
Other popular segmentation approaches include k-means, mean-shift, region splitting and watershed \cite{szeliski2010computer}.

The methods introduced so far extract hand-crafted features which are classified or clustered with a machine learning algorithm.
Deep learning methods, particularly CNNs, have largely outperformed most of these methods
in classification and segmentation tasks. Deep learning approaches enable replacement of hand-crafted descriptors by
trainable filters combined in a cascade of layers, learning multiple levels of abstraction.
CNNs are excellent image recognition networks adapted for texture classification in \cite{andrearczyk2016using,cimpoi2016deep},
in which the overall shape analysis is discarded
as it is irrelevant in texture analysis as opposed to object recognition.
The response to deep filters are extracted from the network for training a classifier in \cite{cimpoi2016deep}
whereas the network is trained end-to-end in \cite{andrearczyk2016using,andrearczyk2016deep} with back-propagation.
An attempt to texture semantic segmentation is also proposed in \cite{cimpoi2016deep} by using a region proposal based on
low level image cues and classifying these regions using the proposed deep descriptors extracted from the CNN.
FCNs are introduced in \cite{long2015fully} to perform pixel-wise semantic segmentation.
These networks are trained end-to-end similarly to CNNs except that an error is calculated for each label.
To obtain a pixelwise classification, fully-connected layers are first replaced by $1\times 1$ convolution layers to
allow arbitrary input sizes.
The output feature maps of these $1\times 1$ convolution layers are smaller than the number of input pixels 
and are therefore upsampled by deconvolution and interpolation to get back to a pixelwise representation.
Following the idea of hypercolumn in \cite{hariharan2015hypercolumns}, information obtained at multiple depths in the network is combined
by using connections to skip certain layers. 
A skip connection combines the final prediction layer with earlier layers of finer strides.
These connections involve $1\times 1$ convolutions for dimensionality reduction of the finer feature maps as well as appropriate deconvolution and subsampling for summation.
In this way, the local information (``where") in intermediate layers is combined to the global information (``what") in deep layers.
This approach allows to extract deep features of high complexity while maintaining a locality information crucial for segmentation across boundaries.

Finally, other deep learning methods have been introduced for the segmentation of various types of images
with different architectures (e.g. U-Net \cite{ronneberger2015u}), deeper networks
and different learning methods (e.g. generative adversarial network \cite{luc2016semantic}).
While these latest advances are of high interest and may be considered in future research,
they are out of the scope of this paper as we base this work on FCNs.

\section{Method description} \label{sec:methodDescription}
This section describes our FCN architecture for texture segmentation as well as a simple segmentation refinement method.

\subsection{Network architecture}
\label{subsec:NetArch}
We adapt a fully convolutional architecture for texture segmentation\footnote{The network architecture implemented with Caffe \cite{jia2014caffe} and more details will be provided.}.
As mentioned in \cite{andrearczyk2016using,cimpoi2016deep}, the overall shape analysis, required in classic object recognition tasks,
can be discarded for the analysis of texture.
In a convolutional network, the semantic or high level of abstraction information (global shape) is contained in the deepest convolution layers and fully connected layers
(implemented with $1\times 1$ convolutions).
For example in the VGG-16 network \cite{simonyan2014very}
trained on Imagenet \cite{deng2009imagenet}, the neurons in the last convolution layer respond to complex shapes in the input image such as faces, cars, persons etc.
Therefore we develop an architecture based on the FCN8 network developed in \cite{long2015fully}
with less convolution layers. We use four sets of convolution layers instead of five as the texture patterns complexity
is less than the shape complexity of objects segmented in \cite{long2015fully}.

\begin{figure}[!t]
\centering
\includegraphics[scale=.65]{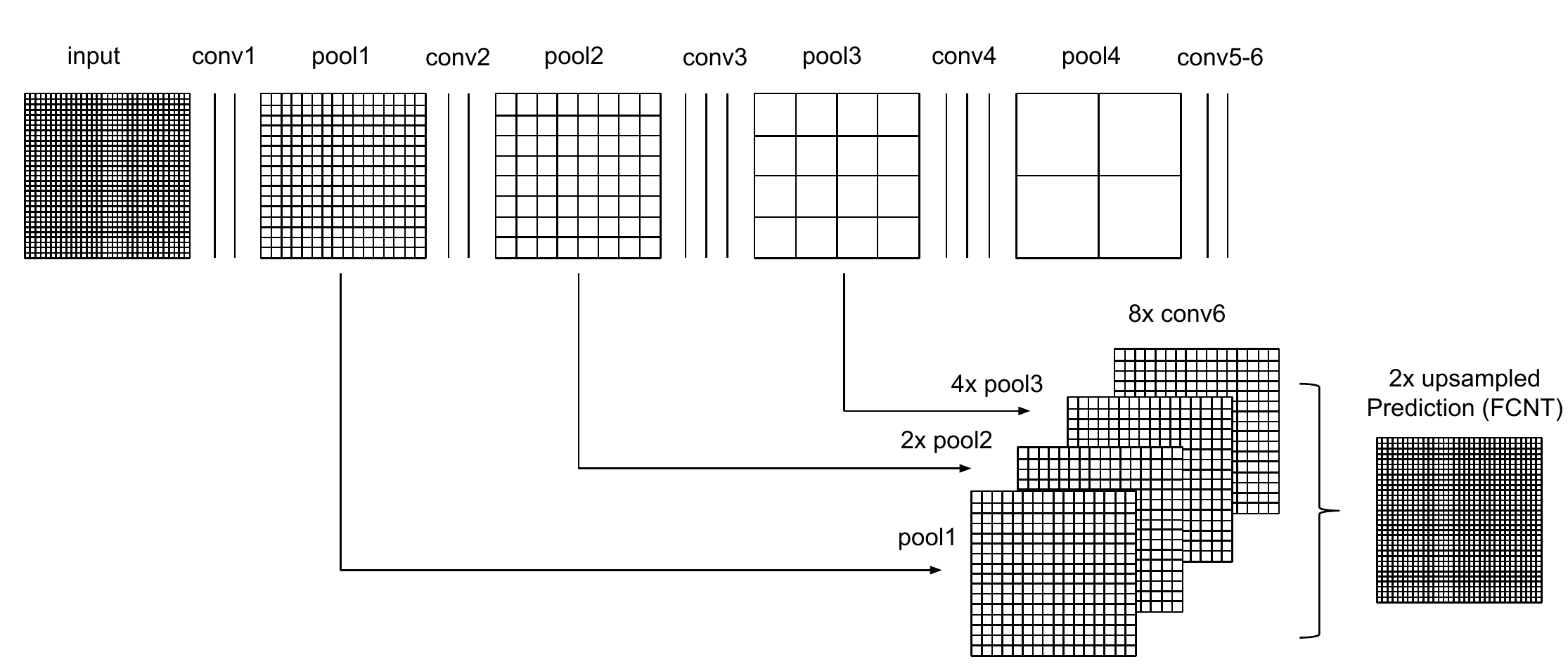}
  \caption{Our FCNT architecture based on \cite{long2015fully}. The grids reveal the relative spatial coarseness of pooling and prediction layers.
  Convolution layers are shown as vertical lines. Skip connections, represented by arrows allow to combine local information
  from early layers with more global information extracted by deeper layers (conv6).
  The upsampling is performed by successive deconvolution operations.
  Note that conv5-6 are the equivalent of fully-connected layers in classic CNN architectures. Diagram adapted from \cite{long2015fully}.
  }\label{fig:diagram}
\end{figure}

Segmentation difficulties can occur across texture boundaries due to downsampling in deep networks, resulting in large receptive fields with high
abstraction and a loss of local information. 
This issue can be addressed by using skip connections \cite{long2015fully} in FCNs in order to combine local and global information in the deconvolution layers.
This is very relevant in texture segmentation as it does not only improve the boundaries segmentation but also provides
high frequency details which often carry useful information about the textures.
Where the FCN8 fuses the outputs of the third and fourth convolution blocks (Conv3 and Conv4) using skip layers,
we aggregate the local information of the outputs of the first, second and third convolution blocks instead.
This architecture that we refer to as FCNT is shown in Figure \ref{fig:diagram}.

\subsection{Refinement of segmented regions}\label{subsec:refinement}
As explained in the previous section, we use local information from shallow layers to significantly improve the boundary segmentation.
However, boundary pixels remain a difficulty and the major cause of misclassification of the FCNT.
For this reason, we develop a segmentation refinement method based on the output of the network.
In the classic inference phase of a FCN, each pixel is assigned a prediction vector of size equal to the number of training classes.
The segmentation result is obtained by assigning each pixel to the class which corresponds to the highest score in the prediction vector.
However, more information is available as each class is assigned a score in these vectors.
We make use of this extra information in a basic refinement method to improve the segmentation results
and to obtain a single texture region per class in experiments B and C.
To do so, $N$ largest patches are first isolated and filled (step 1), where $N$ is the number of texture classes in the segmented image to refine.
A patch refers to a connected region of individual class label.
As shown in Figure \ref{fig:expB}.f, there are generally more than one patch per texture label before refinement.
In an iterative process, the small isolated regions (pixels not part of any largest patch) are then assigned the second best prediction of the network
to attempt to merge them with larger patches (step 2).
Successively, the remaining isolated regions are assigned to class labels with incrementally lower probability scores in the output vectors (steps 4, 5, etc.)
The refinement stops when a segmentation with a single patch per class is obtained.
The pseudo code for this approach is given as follows. 
\\ \hspace*{0.5cm} 1) Largest patches
\\ \hspace*{0.5cm} while($\bar{N}\neq N$):
    \\ \hspace*{1cm} 2) Relabel 1, largest patches
    \\ \hspace*{1cm} 3) Relabel 2, largest patches
    \\ \hspace*{1cm} 4) if $im_3^{t-1}==im_3^{t}$: relabel 3, largest patches;
    \\ \hspace*{1.4cm} else: back to 2)
    \\ \hspace*{1cm} 5) if $im_4^{t-1}==im_4^{t}$: relabel 4, largest patches; 
    \\ \hspace*{1.4cm} else: back to 2)
    \\ \hspace*{1cm} 6) if $im_5^{t-1}==im_5^{t}$: relabel 5, largest patches; 
    \\ \hspace*{1.4cm} else: back to 2)
    \\ \hspace*{1cm} etc.
\\ \hspace*{0.5cm} end while,
\\where $\bar{N}$ and $N$ are respectively the number of patches and the number of texture classes.
``Largest patches" means that we take the largest patches of each class and fill them,
i.e. the labels of the small regions fully contained in the N largest patches are changed to the patches' labels.
``Relabel $x$" refers to the relabelling of all the pixels which are not part of the largest patches to the $x^{th}$ best class prediction of the network at these pixel locations.
Finally $im_X^{t}$ is the output of step X of the pseudo code at the current loop iteration.
An example of refinement result is shown in Figure \ref{fig:expB}.f and \ref{fig:expB}.g.

\section{Experiments}
In this section we evaluate the developed method on a set of supervised and unsupervised experiments.
We conduct three types of experiments (A, B and C) as follows.
\subsection{Experiment A: Supervised training with multiple training images per class}\label{subsec:experimentA}
This task is relatively similar to a classic FCN training method as multiple training instances of the same class are used.
The intra-class variation forces the network to learn certain invariances to scale, orientation, and even types of textures.
The major differences are the texture nature of the images and the fact that the training images are not segmented
(i.e. homogeneous texture across the entire image).

\subsubsection{Datasets}
For this experiment, we develop two datasets as follows\footnote{Please contact the authors to obtain these datasets.}.
For the \textbf{kth-seg} dataset, we use images from the kth-tips-2b dataset \cite{hayman2004significance} which contains 11 classes of texture images.
Each class is made of four groups and each group contains images at 9 different scales.
As we use images at scale 10, each group in a particular class contains 12 images for a total of 528 images.
The evaluation is based on the idea of 4-fold cross-validation used in \cite{andrearczyk2016using,cimpoi2016deep}. Each group is used once to create testing
images of multiple texture regions while the remaining groups are used for training.
To create a test image, we randomly pick a number of images (two to five) from the test group and create a mosaic image with one texture
region from each image (all from different classes).
For each fold, there is a total of 396 training images and 80 test images.

\begin{figure}[!t]
\centering
\includegraphics[scale=.6]{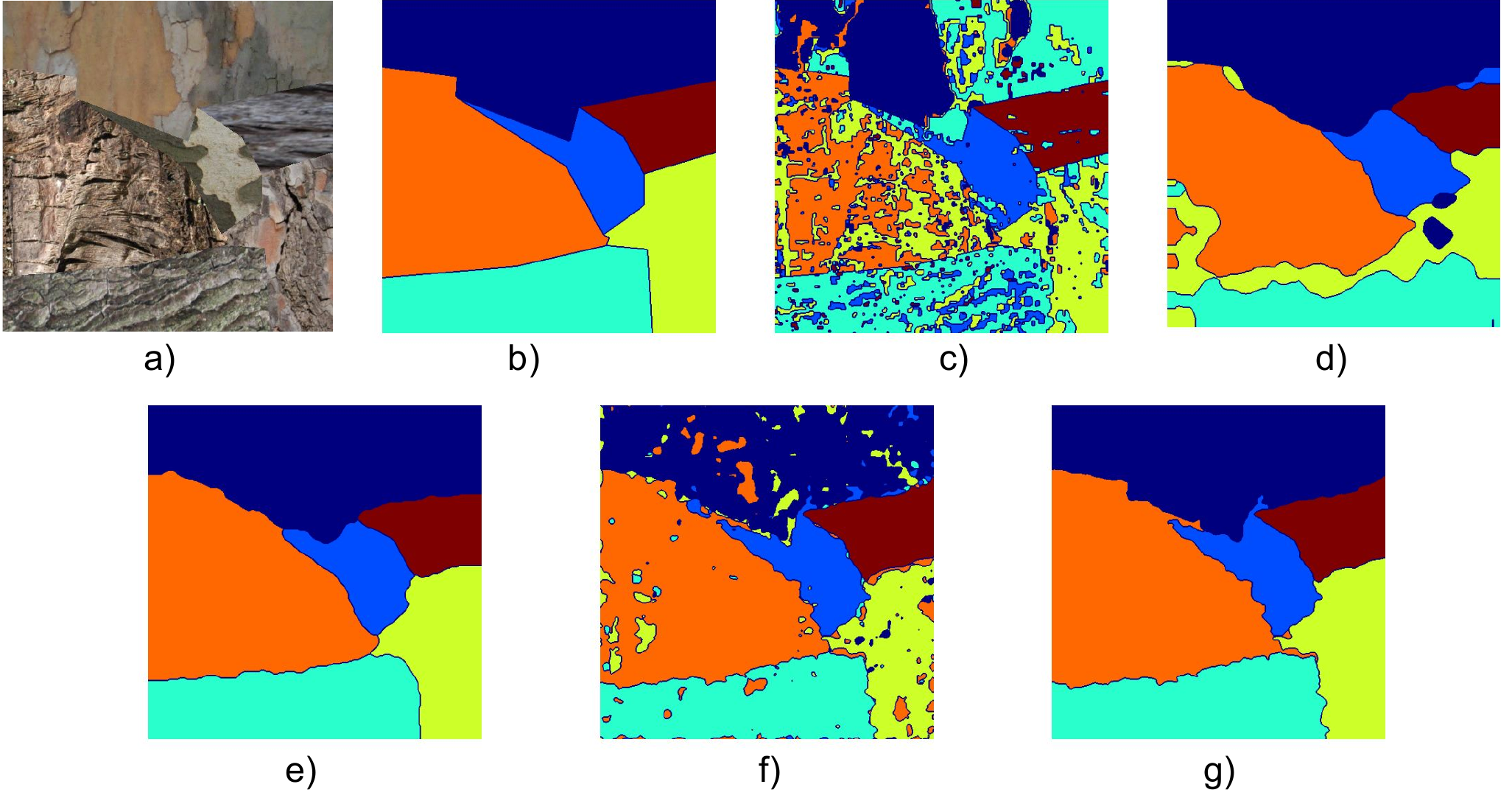}
  \caption{Examples of segmentation with FCNT on the supervised Prague texture segmentation task (experiment B).
  The network is trained with one image per texture segment (six training images in this example).
  a) Input image to segment; b) ground truth segmentation; c) Markov Random Field (MRF) segmentation \cite{kato2001color};
  d) Co-Occurrence Features (COF) segmentation; e) Con-Col segmentation (see Section \ref{subsubsec:results});
  f) FCNT segmentation without refinement (ours); g) FCNT segmentation with refinement (ours).
  Best viewed in color.
  }\label{fig:expB}
\end{figure}
For the \textbf{Kylberg-seg} dataset, we use images from the Kylberg texture classification dataset \cite{Kylberg2011c} which contains 28 classes of 160 images each of size $576\times 576$.
We randomly select one of 12  available orientations for each image similarly to \cite{andrearczyk2016using}.
Half of the dataset is used for training, the rest to create test images.
The images are split into four sub images of size $288\times 288$.
Images from the test set are randomly picked to create test images of size $288\times 288$ with two to five texture regions from different classes, similarly to kth-seg.
There is a total of 8960 training images and 80 test images. Examples of the Kylberg-seg dataset are shown in Figure \ref{fig:expA}.

Note that in such supervised approach with non-segmented images, we could artificially create segmented images at training time to help the network in taking boundary decisions.
However, this would be heavily biased as the artificial segmentation would be very similar to the test segmentation and would not generalize well to real life texture segmentation problems
as the boundary between two real textures can be melted, blurred etc.

\subsubsection{Details of the network}

We initialize the networks (FCN8 and FCNT, see Section \ref{subsec:NetArch}) with the weights learned with FCN8 on the PASCAL VOC 2011 dataset \cite{pascal2011}.
Layers which are not common to both networks (fully-connected and deconvolution layers) are initialized with the Xavier method \cite{glorot2010understanding}. 
The width of the network in the deconvolution layers (number of feature maps in the last fully connected layer and following layers) is equal to the number of
training texture classes, i.e. 11 for kth-seg and 28 for Kylberg-seg.
The networks are trained with stochastic gradient descent similarly to \cite{long2015fully}.
The training data and the amount of information and variation the networks need to learn is relatively large (as compared to experiments B and C).
Therefore we train the networks for 300,000 and 400,000 iterations for the kth-seg and Kylberg-seg experiments respectively.
Neither pre-processing, post-processing nor refinement is applied to demonstrate only the discriminative power of FCNs in texture segmentation with multiple training images per class.

\subsubsection{Results}
\begin{figure}[!t]
\centering
\includegraphics[scale=.65]{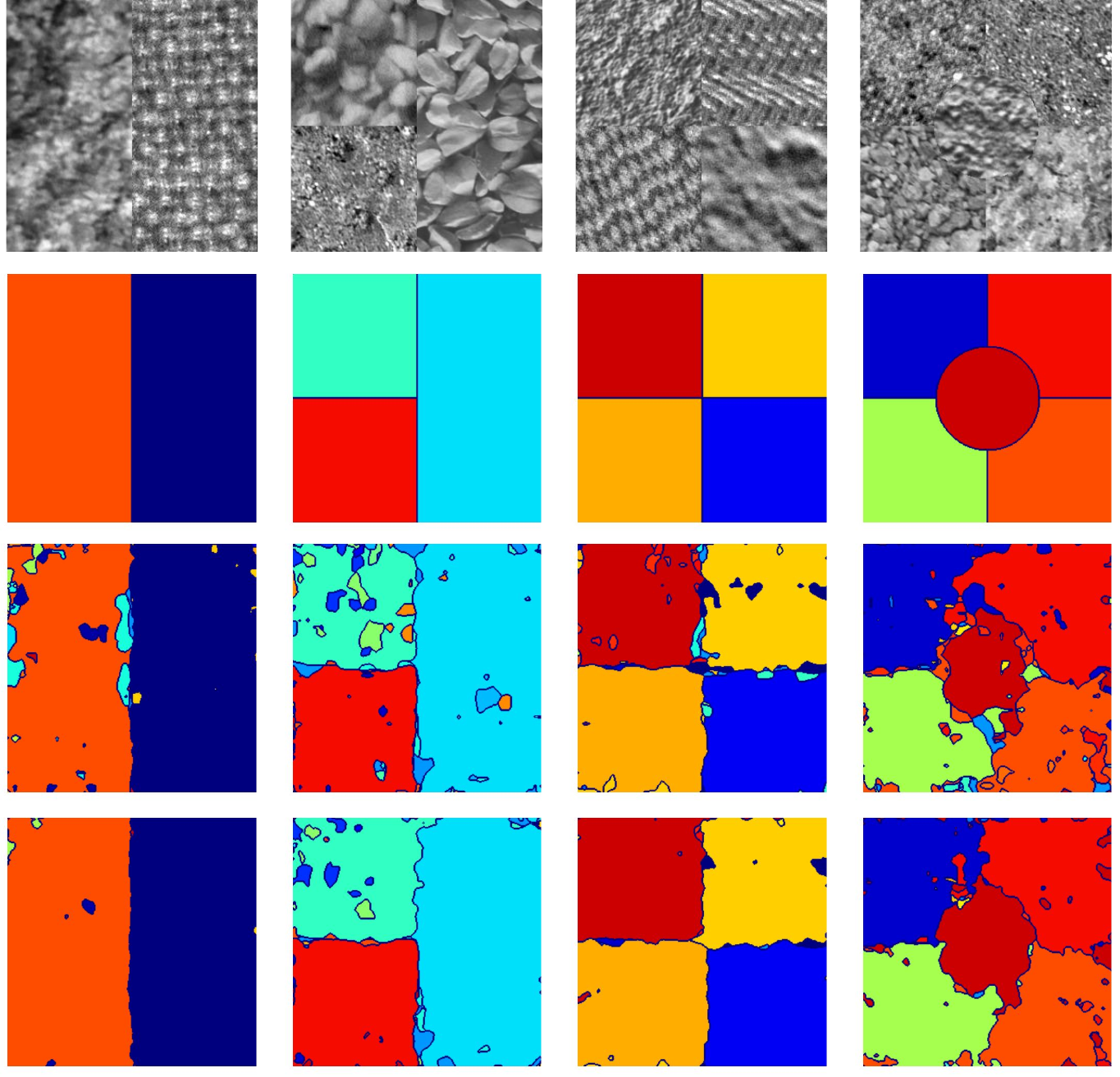}
  \caption{Examples of segmentation with FCN8 \cite{long2015fully} and our FCNT architecture on experiment A (Kylberg-seg).
  First row: input test image to segment. Second row: ground truth segmentation.
  Third row: FCN8 segmentation. Fourth row: FCNT segmentation.
  Best viewed in color.
  }\label{fig:expA}
\end{figure}

In this experiment, we compare our framework using FCN8 developed in \cite{long2015fully} and our FCNT network.
The results are summarized in Table \ref{tab:results_a} as the average correct assignment of pixels.
Figure \ref{fig:expA} illustrates segmentation result examples with two to five texture regions per image.
Note that the segmentation results are directly the output of the network which learned from non-segmented images to recognize multiple classes
(11 for kth-seg and 28 for Kylberg-seg) with challenging inter-class similarities and a high intra-class variation for kth-seg.
The FCNT outperforms FCN8 on the developed texture segmentation datasets (except kth-5) which demonstrates the importance of using local information and discarding the overall shape analysis.
However, the best architecture might depend on the application, e.g. the complexity of the texture patterns, the scale variation, the number of texture regions
as well as the desired results (e.g. precise boundary decision, minimum over-segmentation, etc.).

\begin{table*}[!t]
\caption{Average correct pixel assignment (CO) of the proposed approach with FCN8 and FCNT networks on the developed kth-seg and Kylberg-seg datasets (experiment A).
The numbers next to the datasets (e.g. kth-X) represent the number of texture regions per image.}\label{tab:results_a}
\centering
\begin{tabular}{ |p{1.4cm}|p{1.3cm}|p{1.3cm}|p{1.3cm}|p{1.3cm}|p{1.6cm}|p{1.6cm}|p{1.6cm}|p{1.6cm}| }
\hline
	Datasets    & kth-2     & kth-3         & kth-4     & kth-5     & Kylberg-2     & Kylberg-3     & Kylberg-4     & Kylberg-5     \\ \hline
	FCN8        & 67.93     & 72.60         & 67.76     & 64.98     & 85.94         & 79.13         & 77.43         & 76.33         \\ \hline
	FCNT        & 68.70     & 73.05         & 68.31     & 63.60     & 88.81         & 80.59         & 79.11         & 76.80         \\ \hline
\end{tabular}
\end{table*}

\subsection{Experiment B: Supervised training with single training image per class}\label{subsec:experimentB}
This experiment is conducted using the Prague texture segmentation benchmark developed in \cite{haindl2008texture}
which enables to test and compare algorithms on a range of supervised and unsupervised texture segmentation tasks.

\subsubsection{Dataset}
In this experiment, we use the supervised dataset of the Prague texture segmentation benchmark \cite{haindl2008texture} with 20 test images.
Each test image contains a number of texture regions (of different classes) ranging from 3 to 12 and a single training image is provided for each class
(i.e. N training images for a test image containing N texture regions).
Although different images are used for the training and testing sets, the variation between training and testing images is limited due to their visual similarity.

\subsubsection{Details of the network}
As in experiment A, we initialize the FCNT (see Section \ref{subsec:NetArch}) with the weights learned with FCN8 on the PASCAL VOC 2011 dataset \cite{pascal2011}
and layers which are not common to both networks are initialized with the Xavier method.
The width of the network in the deconvolution layers is again equal to the number of
training texture classes which is given for each test image  by the number of training image (3 to 12), and in turn is equal to the number of ground truth regions in
the test image.
We train for 5000 iterations similarly to \cite{long2015fully} although the results are stable to small changes of this number with a fast learning and negligible overfit.

\subsubsection{Results}\label{subsubsec:results}
For this experiment, we compare our method to three algorithms whose results are provided on the Prague texture segmentation website \cite{prague}.
The algorithm we refer to as MRF is an implementation of the method developed in \cite{kato2001color},
an image segmentation algorithm using a Markov Random Field pixel classification model.
Little information is provided about the second algorithm, referred to as COF as it uses co-occurrence features with a nearest neighbor classifier. 
Finally, no information is provided regarding the best reported results on this supervised dataset. We refer to this state of the art as Con-Col as named in \cite{prague}.
The results for this experiment are summarized in Table \ref{tab:results_b} in the form of various segmentation evaluations including region based (e.g. correct segmentation CS),
pixel-wise (e.g. correct assignment CO) and consistency (e.g. global consistency error GCE) measures.
For more information on these performance measures, the reader should refer to \cite{haindl2008texture,prague}.
The proposed FCNT (with refinement) largely outperforms the best results from the literature and largely benefits from the segmentation refinement method
(see Section \ref{subsec:refinement}).
Figure \ref{fig:expB} shows examples of segmentation results of the three algorithms from the literature as well as our method with and without segmentation refinement.

\begin{table*}[!t]
\caption{Results of experiment B on the Prague supervised dataset (normal size) and comparison with the state of the art.
The results of our approach before segmentation refinement are referred to as FCNT-NR (no refinement).
The performance measures are described in \cite{haindl2008texture,prague}.
Up arrows in the first column indicate that larger values correspond to better results and down arrows the opposite.
Results marked with * indicate that no publication is currently known. } \label{tab:results_b}
\centering
\resizebox{0.7\textwidth}{!}{
\begin{tabular}{ |p{1.4cm}|p{1.2cm}|p{1.2cm}|p{1.9cm}|p{2.4cm}|p{2cm}| }
\hline
	Method              & MRF*      & COF*          & Con-Col*    & FCNT-NR    & FCNT              \\ \hline
     \multicolumn{6}{|c|}{Region-based measures}\\ \hline
	$\uparrow$ CS       & 46.11     & 52.48         & 84.57         & 87.52                 & \textbf{96.01}    \\ \hline
	$\downarrow$ OS     & 0.81      & \textbf{0.00} & \textbf{0.00} & \textbf{0.00}         & 1.56              \\ \hline
	$\downarrow$ US     & 4.18      & 1.94          & 1.70          & \textbf{0.00}         & 1.20              \\ \hline
	$\downarrow$ ME     & 44.82     & 41.55         & 9.50          & 6.70                  & \textbf{0.78}     \\ \hline
	$\downarrow$ NE     & 45.29     & 40.97         & 10.22         & 6.90                  & \textbf{0.89}     \\ \hline
     \multicolumn{6}{|c|}{Pixel-wise measures}\\ \hline
	$\uparrow$ CA       & 65.42     & 67.01         & 86.21         & 87.08                 & \textbf{93.95}    \\ \hline
	$\uparrow$ CO       & 76.19     & 77.86         & 92.02         & 92.61                 & \textbf{96.73}    \\ \hline
	$\uparrow$ CC       & 80.30     & 78.34         & 92.68         & 93.26                 & \textbf{97.02}    \\ \hline
	$\uparrow$ EA       & 75.40     & 76.21         & 91.72         & 92.68                 & \textbf{96.68}    \\ \hline
	$\uparrow$ MS       & 64.29     & 66.79         & 88.03         & 88.92                 & \textbf{95.10}    \\ \hline
	$\uparrow$ CI       & 76.69     & 77.05         & 92.02         & 92.81                 & \textbf{96.77}    \\ \hline
	$\downarrow$ O      & 14.52     & 20.74         & 7.00          & 7.49                  & \textbf{2.72}     \\ \hline
	$\downarrow$ C      & 16.77     & 22.10         & 5.34          & 6.16                  & \textbf{2.29}     \\ \hline
	$\downarrow$ I.     & 23.81     & 22.14         & 7.98          & 7.39                  & \textbf{3.27}     \\ \hline
	$\downarrow$ II.    & 4.82      & 4.40          & 1.70          & 1.49                  & \textbf{0.68}     \\ \hline
	$\downarrow$ RM     & 6.43      & 4.47          & 2.08          & 1.38                  & \textbf{0.86}     \\ \hline
     \multicolumn{6}{|c|}{Consistency measures}\\ \hline
	$\downarrow$ GCE    & 25.79     & 23.94         & 11.76         & 12.54                 & \textbf{5.55}     \\ \hline
	$\downarrow$ LCE    & 20.68     & 19.69         & 8.61          & 9.94                  & \textbf{3.75}     \\ \hline
\end{tabular}
}
\end{table*}

\subsection{Experiment C: Unsupervised training}\label{subsec:experimentC}

\subsubsection{Dataset}
In the last experiment, we use the Prague unsupervised texture segmentation dataset (ICPR 2014 contest) \cite{haindl2008texture}.
It contains 80 test images (which include the 20 test images of experiment B) without any training sample.
Each test image contains 3 to 12 texture regions which must all be segmented as individual textures. 

\subsubsection{Pre-segmentation}
While this experiment is unsupervised, labels are needed to train the network and so
we need to obtain a rough pre-segmentation map from the test images.
The idea is that the FCNT can learn from little training texture data and that textures are homogeneous across the regions of the test images.
Therefore we can train the network from patches of texture regions roughly segmented in an unsupervised manner from the test images.
Two methods are tested to obtain a pre-segmentation map.

First a simple k-means clustering method is applied on the output of a pre-trained FCNT with the image to segment as input.
To that end, the input image to the pre-trained network is downsampled (with  bilinear interpolation) by a factor of four in order to obtain more homogeneous label regions and a fast clustering.
A k-means algorithm with known number of clusters is run on the output of the FCNT. The k-means clustering is performed in a feature space of dimensionality
equal to the number of classes in the pre-trained FCNT network (11 in this case as it is pre-trained on kth-tips-2b).
Note that in the ICPR 2014 contest \cite{haindl2008texture}, the number of clusters is unknown.
This experiment is designed to demonstrate the power of the FCNT
in a texture segmentation with rough pre-segmentation and not to evaluate such pre-segmentation methods.
We therefore use a known number of clusters as it provides a simple pre-segmentation.
We leave the problem of obtaining more accurate pre-segmentation labels from the network with an unknown number of textures for future work.
The obtained label image is then upsampled by the same factor used to downsample the input image.
The largest connected label region for each class is filled and used for training while other small isolated regions and dilated borders between regions are not used.

In a second approach, we use shallow segmentation algorithms from the literature to obtain the pre-segmentation labels.
We use the segmentation results (provided in \cite{prague}) from four algorithms described as follow.
Factorization based texture Segmentation (FSEG) \cite{yuan2015factorization}
uses  local distribution of filter responses (local spectral histograms) to construct a feature matrix.
This matrix is factorized based on singular value decomposition into two matrices, one containing representative features,
the other one containing the weights used for linear combination of the representative features at each pixel location.
These weights indicate the segmentation map result.
The variational multi-phase segmentation framework (PCA-MS) developed in \cite{mevenkamp2016variational} also 
uses a filter bank approach. It performs a dimensionality reduction (principal component analysis) of the extracted local spectral histograms and an energy
(Mumford-Shah) minimization segmentation.
The Priority Multi-Class Flooding Algorithm (VRA-PMCFA) developed in \cite{voting,panagiotakis2011natural}
involves a voting selection of a parameterized number of representative blocks of pixels based on wavelet distributions, 
followed by an iterative computation of segmentation maps by clustering, flooding and region merging.
Finally, the last results appear as the state of the art on the unsupervised Prague dataset \cite{prague}, although
no information is provided about the algorithm.
We refer to this algorithm as MK in reference to the author's name (Martin Kierche).

\subsubsection{Details of the network}
When using a k-means pre-segmentation, the FCNT is pre-trained on kth-tips-2b (only for the pre-segmentation) to obtain a first output without finetuning.

For training the network with the pre-segmentation map (from k-means and other algorithms), the FCNT is initialized similarly to experiments A and B.
The test image itself is fed as training data with training labels being the pre-segmentation map.
The width of the network in the deconvolution layers is again equal to the number of
training texture classes, given for each test image by the number of classes present in the pre-segmentation map.

As we use patches of the test image for training with rough pre-segmentation labels, we do not want the network to overfit.
We assume that a majority of pixels in the pre-segmentation map are correctly classified and that the FCNT will first learn to segment
these pixels which form a region of homogeneous texture.
Misclassified pixels in the pre-segmentation map should exhibit a different texture.
Yet, if it overfits, the network will learn a representation that merges these outliers and the correct regions into the same class.
This is due to both the plasticity of the network and the small amount of training data.
In order to avoid overfitting, we evaluate the output of the network on the test image during training.
We stop the training 60 iterations after all the texture classes present in the pre-segmentation map
are detected in the output of the trained network. In other words when at least one pixel is assigned to each class in the output of the network.
This early stop method avoids overfitting the pre-segmentation training labels.
Note that we limit the number of iterations to 400 if the previous condition is not reached.

\subsubsection{Results}

\begin{figure*}[!t]
\centering
\includegraphics[scale=.5]{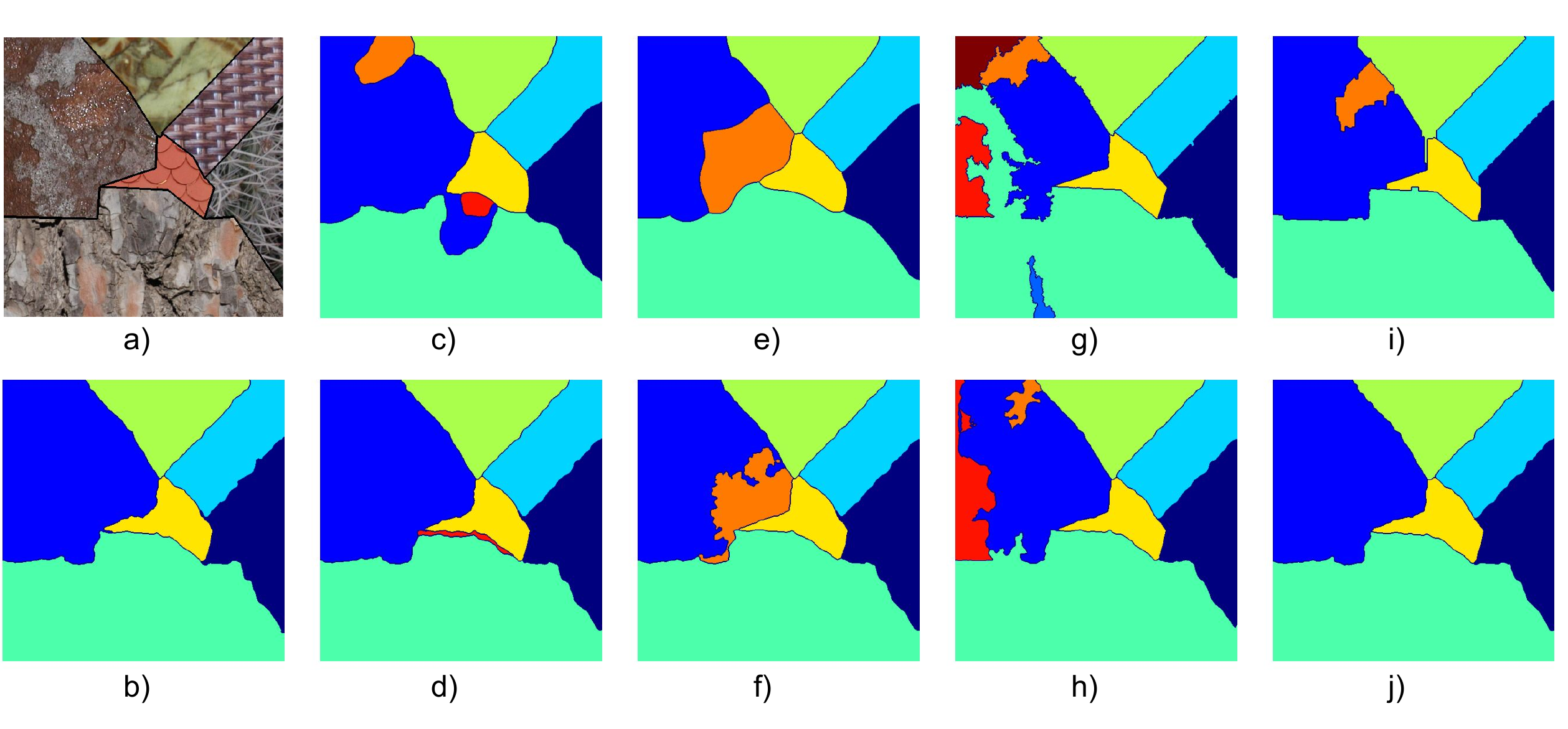}
  \caption{Examples of segmentation results with different methods on the Prague unsupervised dataset.
  a) Input image with superimposed ground truth boundaries; b) FCNT segmentation with k-means training (ours); c) FSEG segmentation
  d) FCNT segmentation with FSEG pre-segmentation (ours); e) PCA-MS segmentation; f) FCNT segmentation with PCA-MS pre-segmentation (ours);
  f) PMCFA segmentation; g) FCNT segmentation with PMCFA training (ours);
  h) MK segmentation; i) FCNT segmentation with MK pre-segmentation (ours); 
  Best viewed in color.
  }\label{fig:expC}
\end{figure*}

The results are reported in Table \ref{tab:results_c} with the same measures as in experiment B.
Images of segmentation results obtained with the shallow algorithms as well as our FCNT approach with different pre-segmentation methods
are shown in Figure \ref{fig:expC}.
Our method used with existing pre-segmentation labels (FSEG, PCA-MS, VRA-PMCFA and MK) consistently improves the results
from the segmentation of these shallow algorithms.
The FCNT trained with labels from the MK segmentation largely outperforms the state of the art.
When compared to the literature, it obtains the best score on 12 out of 18 measurements and second best score on four others.

\begin{table*}[!t]
\caption{Results of our FCNT approach with various pre-segmentation methods on the Prague unsupervised dataset (large size) and comparison with the state of the art.
Results for FSEG, PCA-MS VRA-PMCFA and MK are reported as given on the Prague texture dataset website \cite{prague}.
Results marked with * indicate that no publication is known at the time of writing.} \label{tab:results_c}
\centering
\resizebox{1\textwidth}{!}{
\begin{tabular}{ |p{2cm}|p{1.32cm}||p{0.9cm}|p{0.9cm}||p{1.45cm}|p{1.45cm}||p{2.5cm}|p{2.25cm}||p{0.8cm}|p{0.95cm}| }
\hline
	 Method             & FCNT          & FSEG              & FCNT              & PCA-MS            & FCNT              & VRA-PMCFA*        & FCNT              & MK*               & FCNT              \\ \hline
	FCNT labels         & k-means       & -                 & FSEG              & -                 & PCA-MS            & -                 & VRA-PMCFA         & -                 & MK                \\ \hline
     \multicolumn{10}{|c|}{Region-based measures}\\ \hline
    $\uparrow   $ CS    & 62.59         & 69.24             & 71.90             & 72.27             & 75.00             & 75.14             & 75.62             & 77.73             & \textbf{79.34}    \\ \hline
	$\downarrow $ OS    & 13.02         & 12.28             & \textbf{9.82}     & 18.33             & 17.09             & 12.13             & 11.51             & 15.92             & 13.67             \\ \hline
	$\downarrow $ US    & 16.31         & 17.03             & 19.86             & 9.41              & 9.69              & 9.85              & 10.17             & 6.31              & \textbf{6.25}     \\ \hline
	$\downarrow $ ME    & 13.98         & 7.71              & 4.73              & 4.19              & \textbf{3.62}     & 4.38              & 4.76              & 3.93              & 3.80              \\ \hline
	$\downarrow $ NE    & 14.09         & 6.89              & 4.26              & 3.92              & \textbf{3.32}     & 4.37              & 4.66              & 3.92              & 3.80              \\ \hline
     \multicolumn{10}{|c|}{Pixel-wise measures measures}\\ \hline
	$\uparrow   $ CA    & 74.58         & 76.32             & 78.67             & 81.13             & 83.02             & 83.45             & 83.77             & 82.80             & \textbf{84.17}    \\ \hline
	$\uparrow   $ CO    & 82.79         & 84.05             & 85.62             & 85.96             & 87.41             & 88.12             & \textbf{88.54}    & 86.89             & 87.97             \\ \hline
	$\uparrow   $ CC    & 83.90         & 84.15             & 84.35             & 91.24             & 91.77             & 90.73             & 90.51             & 93.65             & \textbf{94.15}    \\ \hline
	$\uparrow   $ EA    & 81.42         & 82.53             & 83.71             & 87.08             & 88.20             & 88.07             & 88.23             & 88.03             & \textbf{88.97}    \\ \hline
	$\uparrow   $ MS    & 74.18         & 77.11             & 79.18             & 81.84             & 83.61             & 83.92             & 84.28             & 83.98             & \textbf{85.23}    \\ \hline
	$\uparrow   $ CI    & 82.30         & 83.26             & 84.32             & 87.81             & 88.87             & 88.72             & 88.80             & 89.03             & \textbf{89.91}    \\ \hline
	$\downarrow $ O     & 10.43         & 11.66             & 9.99              & 7.25              & 5.34              & \textbf{4.51}     & 4.76              & 7.68              & 6.47              \\ \hline
	$\downarrow $ C     & 12.45         & 11.71             & 8.46              & 6.44              & \textbf{5.53}     & 8.89              & 7.81              & 24.24             & 22.88             \\ \hline
	$\downarrow $ I.    & 17.21         & 15.95             & 14.38             & 14.04             & 12.59             & 11.88             & \textbf{11.46}    & 13.11             & 12.03             \\ \hline
	$\downarrow $ II.   & 3.56          & 3.29              & 3.11              & 1.59              & 1.47              & 1.48              & 1.48              & 1.50              & \textbf{1.42}     \\ \hline
	$\downarrow $ RM    & 5.21          & 5.00              & 5.11              & 4.45              & 4.25              & 3.75              & 3.71              & 3.27              & \textbf{3.12}     \\ \hline
     \multicolumn{10}{|c|}{Consistency measures}\\ \hline
	$\downarrow $ GCE   & 14.21         & 10.75             & 7.62              & 8.33              & 6.75              & 6.55              & \textbf{6.39}     & 7.40              & 6.46              \\ \hline
	$\downarrow $ LCE   & 8.17          & 7.52              & 4.97              & 5.61              & 4.10              & \textbf{3.90}     & 3.97              & 5.62              & 4.75              \\ \hline
\end{tabular}
}
\end{table*}

Note that these results are obtained on the Prague dataset referred to as 'large' in \cite{haindl2008texture} (80 test images)
whereas the supervised results are given in Table \ref{tab:results_b} for the 'normal' Prague dataset (20 test images)
in order to compare to the state of the art.
A comparison is however possible as the results on the normal and large supervised datasets are very similar and the large dataset is an extended version of the normal one.
For comparison, the supervised method obtains 95.64\% correct segmentation (CS) on the large dataset (vs. 96.01\% on the normal dataset);
as expected, significantly better than the unsupervised best results (79.34\%).

\section{Conclusion}
This work has shown the power of FCNs in the segmentation of textures, with features of multiple complexity trained end-to-end to recognize and segment texture regions.
We developed a fully convolutional architecture designed for texture segmentation by combining shallow features encoding local high frequency information
and deeper features containing more abstract information.
We explored several tasks including supervised and unsupervised segmentation.
In particular, we showed that the developed FCN architecture can learn to perform semantic segmentation from classic texture recognition datasets with
non-segmented images (kth-tips-2b and Kylberg).
We have also shown that it can be trained (finetuned) with a very small amount of training data (single image per class) and even using texture regions
of the test image itself as training data after a pre-segmentation process.
The proposed approach largely improved the state of the art on the Prague texture segmentation benchmark including supervised and unsupervised datasets.

\bibliographystyle{ieeetr}
\bibliography{refs}

\begin{thebibliography}{10}

\bibitem{haralick1973textural}
R.~M. Haralick, K.~Shanmugam, {\em et~al.}, ``Textural features for image
  classification,'' {\em IEEE Transactions on systems, man, and cybernetics},
  vol.~3, no.~6, pp.~610--621, 1973.

\bibitem{liu2006image}
X.~Liu and D.~Wang, ``{Image and texture segmentation using local spectral
  histograms},'' {\em IEEE Transactions on Image Processing}, vol.~15, no.~10,
  pp.~3066--3077, 2006.

\bibitem{mevenkamp2016variational}
N.~Mevenkamp and B.~Berkels, ``{Variational multi-phase segmentation using
  high-dimensional local features},'' in {\em 2016 IEEE Winter Conference on
  Applications of Computer Vision (WACV)}, pp.~1--9, IEEE, 2016.

\bibitem{yuan2015factorization}
J.~Yuan, D.~Wang, and A.~M. Cheriyadat, ``{Factorization-based texture
  segmentation},'' {\em IEEE Transactions on Image Processing}, vol.~24,
  no.~11, pp.~3488--3497, 2015.

\bibitem{shi2000normalized}
J.~Shi and J.~Malik, ``{Normalized cuts and image segmentation},'' {\em IEEE
  Transactions on pattern analysis and machine intelligence}, vol.~22, no.~8,
  pp.~888--905, 2000.

\bibitem{paragios2002geodesic}
N.~Paragios and R.~Deriche, ``{Geodesic active regions and level set methods
  for supervised texture segmentation},'' {\em International Journal of
  Computer Vision}, vol.~46, no.~3, pp.~223--247, 2002.

\bibitem{andrearczyk2016using}
V.~Andrearczyk and P.~F. Whelan, ``{Using Filter Banks in Convolutional Neural
  Networks for Texture Classification},'' {\em Pattern Recognition Letters},
  vol.~84, pp.~63--69, 2016.

\bibitem{andrearczyk2016deep}
V.~Andrearczyk and P.~F. Whelan, ``Deep learning for biomedical texture image
  analysis,'' in {\em Irish Machine Vision \& Image Processing Conference
  proceedings IMVIP 2016}, 2016.

\bibitem{cimpoi2016deep}
M.~Cimpoi, S.~Maji, I.~Kokkinos, and A.~Vedaldi, ``{Deep filter banks for
  texture recognition, description, and segmentation},'' {\em International
  Journal of Computer Vision}, vol.~118, no.~1, pp.~65--94, 2016.

\bibitem{long2015fully}
J.~Long, E.~Shelhamer, and T.~Darrell, ``{Fully convolutional networks for
  semantic segmentation},'' in {\em Proceedings of the IEEE Conference on
  Computer Vision and Pattern Recognition}, pp.~3431--3440, 2015.

\bibitem{hariharan2015hypercolumns}
B.~Hariharan, P.~Arbel{\'a}ez, R.~Girshick, and J.~Malik, ``{Hypercolumns for
  object segmentation and fine-grained localization},'' in {\em Proceedings of
  the IEEE Conference on Computer Vision and Pattern Recognition},
  pp.~447--456, 2015.

\bibitem{haindl2008texture}
M.~Haindl and S.~Mikes, ``{Texture segmentation benchmark},'' in {\em Pattern
  Recognition, 2008. ICPR 2008. 19th International Conference on}, pp.~1--4,
  IEEE, 2008.

\bibitem{randen1999filtering}
T.~Randen and J.~H. Husoy, ``{Filtering for texture classification: A
  comparative study},'' {\em IEEE Transactions on pattern analysis and machine
  intelligence}, vol.~21, no.~4, pp.~291--310, 1999.

\bibitem{jain1991unsupervised}
A.~K. Jain and F.~Farrokhnia, ``{Unsupervised texture segmentation using Gabor
  filters},'' {\em Pattern recognition}, vol.~24, no.~12, pp.~1167--1186, 1991.

\bibitem{ojala1999unsupervised}
T.~Ojala and M.~Pietik{\"a}inen, ``{Unsupervised texture segmentation using
  feature distributions},'' {\em Pattern Recognition}, vol.~32, no.~3,
  pp.~477--486, 1999.

\bibitem{chen1979segmentation}
P.~C. Chen and T.~Pavlidis, ``{Segmentation by texture using a co-occurrence
  matrix and a split-and-merge algorithm},'' {\em Computer graphics and image
  processing}, vol.~10, no.~2, pp.~172--182, 1979.

\bibitem{arivazhagan2003texture}
S.~Arivazhagan and L.~Ganesan, ``{Texture segmentation using wavelet
  transform},'' {\em Pattern Recognition Letters}, vol.~24, no.~16,
  pp.~3197--3203, 2003.

\bibitem{szeliski2010computer}
R.~Szeliski, {\em {Computer vision: algorithms and applications}}.
\newblock Springer Science \& Business Media, 2010.

\bibitem{ronneberger2015u}
O.~Ronneberger, P.~Fischer, and T.~Brox, ``{U-net: Convolutional networks for
  biomedical image segmentation},'' in {\em International Conference on Medical
  Image Computing and Computer-Assisted Intervention}, pp.~234--241, Springer,
  2015.

\bibitem{luc2016semantic}
P.~Luc, C.~Couprie, S.~Chintala, and J.~Verbeek, ``{Semantic Segmentation using
  Adversarial Networks},'' {\em arXiv preprint arXiv:1611.08408}, 2016.

\bibitem{jia2014caffe}
Y.~Jia, E.~Shelhamer, J.~Donahue, S.~Karayev, J.~Long, R.~Girshick,
  S.~Guadarrama, and T.~Darrell, ``Caffe: Convolutional architecture for fast
  feature embedding,'' {\em arXiv preprint arXiv:1408.5093}, 2014.

\bibitem{simonyan2014very}
K.~Simonyan and A.~Zisserman, ``{Very deep convolutional networks for
  large-scale image recognition},'' {\em arXiv preprint arXiv:1409.1556}, 2014.

\bibitem{deng2009imagenet}
J.~Deng, W.~Dong, R.~Socher, L.-J. Li, K.~Li, and L.~Fei-Fei, ``{Imagenet: A
  large-scale hierarchical image database},'' in {\em Computer Vision and
  Pattern Recognition, 2009. CVPR 2009. IEEE Conference on}, pp.~248--255,
  IEEE, 2009.

\bibitem{hayman2004significance}
E.~Hayman, B.~Caputo, M.~Fritz, and J.-O. Eklundh, ``{On the significance of
  real-world conditions for material classification},'' in {\em Computer
  Vision-ECCV 2004}, pp.~253--266, Springer, 2004.

\bibitem{kato2001color}
Z.~Kato, T.-C. Pong, and J.~C.-M. Lee, ``{Color image segmentation and
  parameter estimation in a markovian framework},'' {\em Pattern Recognition
  Letters}, vol.~22, no.~3, pp.~309--321, 2001.

\bibitem{Kylberg2011c}
G.~Kylberg, ``{The Kylberg Texture Dataset v. 1.0},'' External report (Blue
  series)~35, Centre for Image Analysis, Swedish University of Agricultural
  Sciences and Uppsala University, Uppsala, Sweden, September 2011.

\bibitem{pascal2011}
M.~Everingham, L.~Van~Gool, C.~K.~I. Williams, J.~Winn, and A.~Zisserman, ``The
  {PASCAL} {V}isual {O}bject {C}lasses {C}hallenge 2011 {(VOC2011)}
  {R}esults.''
  http://www.pascal-network.org/challenges/VOC/voc2011/workshop/index.html,
  2016.

\bibitem{glorot2010understanding}
X.~Glorot and Y.~Bengio, ``{Understanding the difficulty of training deep
  feedforward neural networks.},'' in {\em Aistats}, vol.~9, pp.~249--256,
  2010.

\bibitem{prague}
M.~Haindl and S.~Mikes, ``{The Prague Texture Segmentation Datagenerator and
  Benchmark}.'' \url{http://mosaic.utia.cas.cz/index.php?act=view_res}, 2008.
\newblock [Online; accessed 25-January-2017].

\bibitem{voting}
C.~Panagiotakis, ``{Texture Segmentation Based on Voting of Blocks, Bayesian
  Flooding and Region Merging}.''
  \url{https://sites.google.com/site/costaspanagiotakis/research/imagesegmentation},
  2014.
\newblock [Online; accessed 25-January-2017].

\bibitem{panagiotakis2011natural}
C.~Panagiotakis, I.~Grinias, and G.~Tziritas, ``{Natural image segmentation
  based on tree equipartition, bayesian flooding and region merging},'' {\em
  IEEE Transactions on Image Processing}, vol.~20, no.~8, pp.~2276--2287, 2011.

\end{thebibliography}
\end{document}